\documentclass[conference]{IEEEtran}
\IEEEoverridecommandlockouts
\usepackage{cite}
\usepackage{amsmath,amssymb,amsfonts}
\usepackage{algorithmic}
\usepackage{graphicx}
\usepackage{textcomp}
\usepackage{multirow}
\usepackage{subcaption}
\usepackage{amsmath}
\usepackage{xcolor}
\usepackage{comment}
\def\BibTeX{{\rm B\kern-.05em{\sc i\kern-.025em b}\kern-.08em
    T\kern-.1667em\lower.7ex\hbox{E}\kern-.125emX}}
\begin{document}

\title{MREAK : Morphological Retina Keypoint Descriptor\\
}

\author{\IEEEauthorblockN{Himanshu Vaghela}
\IEEEauthorblockA{\textit{Department of Computer Engineering} \\
\textit{D. J. Sanghvi College of Engineering}\\
Mumbai, India \\
himanshuvaghela1998@gmail.com}
\and
\IEEEauthorblockN{Manan Oza}
\IEEEauthorblockA{\textit{Department of Computer Engineering} \\
\textit{D. J. Sanghvi College of Engineering}\\
Mumbai, India \\
manan.oza0001@gmail.com}
\and
\IEEEauthorblockN{Prof. Sudhir Bagul}
\IEEEauthorblockA{\textit{Department of Computer Engineering} \\
\textit{D. J. Sanghvi College of Engineering}\\
Mumbai, India \\
Sudhir.Bagul@djsce.ac.in}
}
\maketitle

\begin{abstract}
A variety of computer vision applications depend on the efficiency of image matching algorithms used. Various descriptors are designed to detect and match features in images. Deployment of this algorithms in mobile applications creates a need for low computation time. Binary descriptors requires less computation time than float-point based descriptors because of the intensity comparison between pairs of sample points and comparing after creating a binary string. In order to decrease time complexity, quality of keypoints matched is often compromised. We propose a keypoint descriptor named Morphological Retina Keypoint Descriptor (MREAK) inspired by the function of human pupil which dilates and constricts responding to the amount of light. By using morphological operators of opening and closing and modifying the retinal sampling pattern accordingly, an increase in the number of accurately matched keypoints is observed. Our results show that matched keypoints are more efficient than FREAK descriptor and requires low computation time than various descriptors like SIFT, BRISK and SURF.

\end{abstract}


\section{Introduction}

Feature detection is used eminently in image mosaicing, object recognition, image classification and many other computer vision applications for improving the precision of results. It has always been a challenge to identify feature points more efficiently. A sufficient increase in number of efficient keypoints leads to a better feature matching which in turn might improve results in various applications.

Many descriptors have been developed in the last few years. Lowe’s SIFT \cite{b1} and SURF \cite{b2} proposed by Bay \textit{$et\;al.$} are significant examples of the family of Histograms of Oriented Gradients (HOG) based descriptors. SIFT \cite{b1} is rotation and scale invariant, but it is mathematically complicated and computationally heavy due to calculation  of gradients of each pixel in the patch which makes it less effective for low powered devices. SURF \cite{b2} is an improvement on SIFT \cite{b1} as it uses a box filter approximation which makes it faster in computation comparatively. These descriptors are conventional methods which uses histograms for orientation and hence are irrational for mobile devices due to high computation time.

\begin{figure}[t!]
  \centering
  \begin{subfigure}[b]{0.34\linewidth}
    \includegraphics[width=\linewidth]{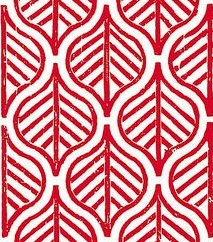}
    \caption{Original}
    \label{fig:ex}
  \end{subfigure}
  \begin{subfigure}[b]{0.34\linewidth}
    \includegraphics[width=\linewidth]{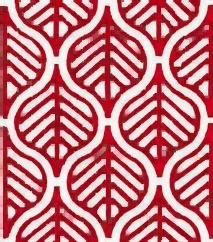}
    \caption{Opening}
    \label{fig:exo}
  \end{subfigure}
  \begin{subfigure}[b]{0.34\linewidth}
    \includegraphics[width=\linewidth]{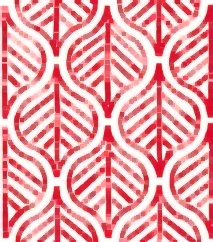}
    \caption{Closing}
    \label{fig:exc}
  \end{subfigure}
  \caption{Morphological operations on an image \cite{b10}}
  \label{fig:morp}
\end{figure}

In recent times, necessity of applications with low computation time is observed due to massive development in the field of mobile technology. To improve performance, binary descriptors were introduced. By comparison of intensity in images, most of the information of a patch is encoded as a binary string. The hamming distance is used as a distance measure between two binary strings which equals the sum of the XOR operation between the two binary strings which reduces computational costs and memory requirements. The Binary Robust Independent Elementary Feature (BRIEF) \cite{b3}, the Oriented Fast and Rotated BRIEF (ORB) \cite{b4},  the Binary Robust Invariant Scalable Keypoints (BRISK) \cite{b5} and the Fast retina keypoints (FREAK) \cite{b6} are good examples. In the next section, we will briefly present these descriptors. In order to decrease computation time in mobile applications, the accuracy of the keypoints detected is ignored. In fact, keypoint detection and matching is an important concept which plays a significant role in various applications.

In this paper, we propose an alternative binary descriptor named MREAK (Morphological Retina Keypoint Descriptor) which is based on the FREAK descriptor and inspired by the pupillary response that varies the size of the pupil in human eye, by adjusting image pixel intensity to increase the number of accurate keypoints and mimic the pupil of human eye. Morphological operations \cite{b8} are used to mimic the function of human pupil which varies the intensity of images in specific areas of an image. Results show that our proposed method detects new feature points more accurately and feature matching is improved due to slight changes in sampling pattern according to required conditions.

\section{Related work}

Several computer vision tasks require matching keypoints across several frames or views. The keypoints need to be detected first. Harris and Stephen in \cite{b7} proposed the Harris corner detector. This detector considers differential of the corner score into account with reference to direction directly, and has been proved to be accurate in distinguishing between edges and corners. 

After keypoint detection, the descriptor is constructed. Float-point based descriptor and binary descriptor are two general types of descriptor. Lowe SIFT \cite{b1} is a state-of-the-art descriptor which belongs to float-point based category and generates robust features which are scale and rotation invariant. A grid is taken around the keypoint and histogram is generated. Finally, a 128-dimensional vector of gradients is taken into consideration. This simple descriptor provides distinctive features, but requires high computational time due to histogram generation. Another descriptor faster than SIFT \cite{b1} is The Speeded up Robust Feature (SURF) by Bay \textit{$et\;al.$} \cite{b2}. SURF uses BLOB detector based on Hessian matrix to detect keypoints. Gaussian weights are applied in all directions for orientation similar as SIFT. 

Float-point based descriptor with more accuracy requires heavy computation time. Due to increasing demand of mobile applications, a demand of low computation time is noted. Binary descriptors use haming distance and XOR operation for matching keypoints and therefore requires less computation time as compared to float-point descriptors. A sampling pattern is formed around the keypoint and a binary string is computed by comparing intensity values of sample points. BRIEF \cite{b3} is a binary descriptor in which there is no specific sampling pattern or any method for orientation calculation. The sampling pairs for forming binary strings are randomly selected. Therefore, BRIEF is less accurate for minute changes in the size or alignment of an image.

ORB \cite{b4} is another example of binary descriptor which has a random sampling pattern but uses a specific method for orientation. Thus, rotation invariance is obtained by calculating moments of the window around keypoint and considering the angle of resultant moment. Furthermore, trained sampling pairs are used which makes it more robust method than BRIEF. BRISK \cite{b5} has a fixed sampling pattern of concentric circles with increasing points away from the centre. Pairs of points are divided in two parts. Long pairs are used for orientation calculation while short pairs are used for forming sampling pattern which makes it better than BRISK and ORB in terms of certain photometric changes. 

FREAK \cite{b6} is a binary descriptor whose sampling pattern is inspired by the distribution of ganglion cells in human retina. The density of points increases moving towards the centre which mimics the human retina. Predefined pairs determine its orientation while sampling points uses coarse-to-fine approach which selects pairs from outer rings followed by pairs from inner rings. In many aspects, FREAK performs better than other binary descriptors.

The binary descriptors are computed on keypoints obtained. While in our approach, more accurate keypoints are detected using morphological operations and sampling pattern of FREAK is modified and implemented on keypoints accordingly to get better matches.  

\section{Human pupil}

\begin{figure}
  \includegraphics[width=\linewidth]{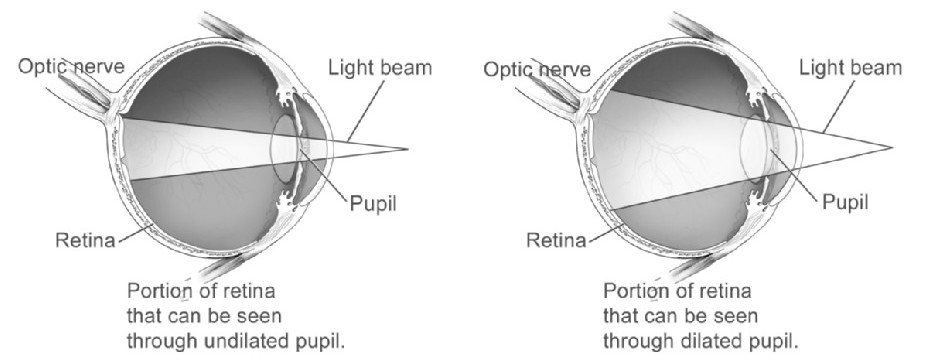}
  \caption{Human pupil \cite{b11}}
  \label{fig:pupil}
\end{figure}

\subsection{Function of pupil}\label{AA}
The function of human pupil is to regulate the amount of light entering human eye. Its shape changes according to the surrounding environment. In darkness, the pupil dilates while it constricts in brightness. This phenomenon decides the amount of light falling on retina as seen in fig.~\ref{fig:pupil} which carries signals to the brain through the optic nerve. By controlling the amount of light entering the eye, improvement in its vision is experienced as the main motive is to view objects clearly in extreme environments.

In computer vision, efforts have been made to mimic the human eye to improve the precision of existing algorithms. In FREAK \cite{b6}, the sampling pattern was inspired from the distribution of ganglion cells over the retina of human eye, which improved the quality of feature points in an image. Our approach is motivated by this function of pupil of varying the intensity of light to increase the number of accurate keypoints detected in an image.

\subsection{Opening and closing}

Image restoration is used to improve numerous imperfections occurring frequently in an image. Morphological image processing is used to extract image components such as boundaries, skeletons, convex hull, etc. This operation can be used in images which are unclear due to the extreme intensity of a group of pixels in a specific part of the image. Dilation and erosion \cite{b8} are operations which are used to expand and diminish brighter areas of an image respectively. While opening and closing \cite{b8} are modifications over the previous ones.  An opening operation is erosion of an image followed by dilation while closing operation is dilation of an image followed by erosion. 

In our approach, opening and closing operations are performed on an image before detecting the keypoints. This morphological operations are used in binary images \cite{b9}. They can be used in grayscale images \cite{b8} which processes a single-channel image. In order to use these operations in coloured image, they are performed in each of the RGB channels. The structuring element or kernel as mentioned in \cite{b8} is used to determine the extent of impact of these operations. A $3\times3$ rectangular structuring element is used in our method. A simple example of opening and closing is shown in Fig.~\ref{fig:morp}.

 \[
   Kernel=
  \left[ {\begin{array}{ccc}
   1 & 1 & 1 \\
   1 & 1 & 1 \\
   1 & 1 & 1 \\
  \end{array} } \right]
\]

\section{The MREAK Descriptor}

\begin{figure}[t!]
  \centering
  \begin{subfigure}[b]{0.48\linewidth}
    \includegraphics[width=\linewidth]{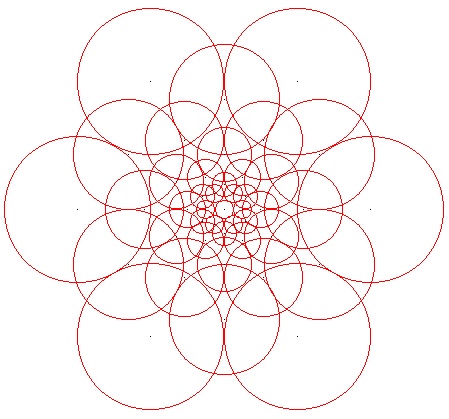}
    \caption{}
    \label{fig:freak}
  \end{subfigure}
  \begin{subfigure}[b]{0.48\linewidth}
    \includegraphics[width=\linewidth]{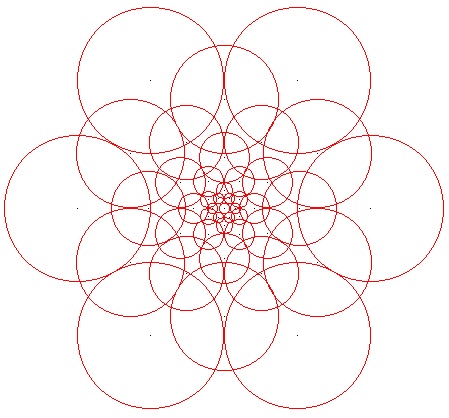}
    \caption{}
    \label{fig:freakopen}
  \end{subfigure}
  \begin{subfigure}[b]{0.48\linewidth}
    \includegraphics[width=\linewidth]{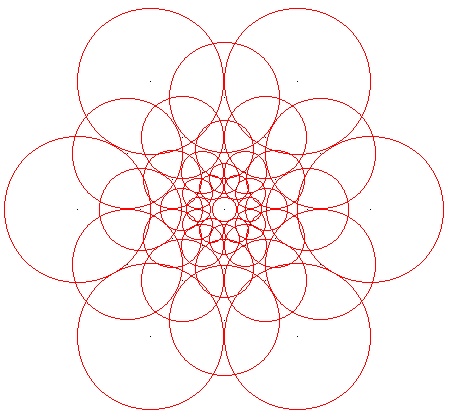}
    \caption{}
    \label{fig:freakclose}
  \end{subfigure}
  \caption{Sampling pattern of (a) FREAK \cite{b6} and proposed (b) Opening and (c) Closing}
  
\end{figure}

\subsection{Sampling pattern}\label{AA}

Sampling pattern around the keypoint plays an important role in matching the keypoints. BRIEF \cite{b3}  does not have a specific sampling pattern while ORB \cite{b4} has a random one. BRISK \cite{b5} was the first descriptor to use a specific pattern to sample its points. FREAK \cite{b6} has a sampling pattern which uses gaussian smoothing over the area around the points and is inspired by the design of human retina.

Our sampling pattern is modification over FREAK by taking morphological operations \cite{b8} like opening and closing into consideration. Fig.~\ref{fig:freak} shows the sampling pattern of FREAK while Fig.~\ref{fig:freakopen} and Fig.~\ref{fig:freakclose} shows sampling pattern of opening and closing respectively proposed by our method. Radius of concentric circles and size of receptive fields are smaller than that of FREAK as we move towards the centre in opening pattern and they are bigger in case of closing pattern. After keypoints are detected in an image, sampling pattern is applied according to the morphological operation used. 

In opening of an image, bright objects are diminished. So, the opening sampling pattern helps us to identify relevant information around the keypoint which is more concentrated towards the centre. While closing of an image expands the bright objects in an image. Therefore, closing sampling pattern highlights relevant information which is concentrated slightly away from the centre. 

This change in sampling pattern improves keypoint detection and constructs a descriptor which enhances the quality of features around the keypoint. Due to increase in number of detected keypoints, image matching efficiency improves. 

\subsection{Orientation}

The focus of our approach is to increase the number of keypoints which leads to better matching. So, for orientation we use same 45 pairs of points as that of FREAK \cite{b6}. This pairs requires less computational memory and time. Orientation method is same as that of FREAK. 

\begin{equation}
O=\frac{1}{M}\sum_{P_0 \in G}(I(P_0^{r1})-I(P_0^{r2}))\frac{P_0^{r1}-P_0^{r2}}{||P_0^{r1}-P_0^{r2}||}
\label{fig:eq1}
\end{equation}

Here, $M$ is the total number of pairs used where $G$ is local gradient and $P_0^{i}$ is the 2D vector of the coordinates of the sample points.

Orientation of points is calculated separately for opened and closed images followed by image matching. By doing this, a separate set of matched keypoints are obtained for opened and closed images. The characteristics of orientation are same as that of FREAK \cite{b6}.

\subsection{The Descriptor}

We compare the intensity values between pairs of receptive fields of sample points with their corresponding Gaussian kernel. A binary string $B$ for $N$ pairs is computed by considering one bit difference of intensities of receptive fields denoted by $T(P_a)$. 

\begin{equation}
B=\sum_{0 \le a < N}2^a T(P_a)
\label{fig:eq2}
\end{equation}

$P_a$ represents a pair of receptive fields whereas $I(P_a^{r1})$ and $I(P_a^{r2})$ denotes intensities of two fields in one pair.  

\begin{equation}	
T(P_a) = \begin{cases} 
       	   1 \quad if (I(P_a^{r1}) > I(P_a^{r2})), \\
       	   0 \quad otherwise,
       	\end{cases}
\label{fig:eq3}
\end{equation}

In order to select pairs from many possible combinations, we use the same algorithm used in FREAK. Following are the steps :

\begin{enumerate}
   \item A descriptor is constructed for each keypoint and it is arranged in form of a matrix with rows representing key points and possible pairs represented by columns. The mean of all columns is calculated.
   \item For a unique feature, high variance would be preferable and it is obtained by mean of 0.5.
   \item The columns are arranged accordingly and best column is selected and rest columns are considered according to their correlation with the selected columns which is less.
\end{enumerate}

Our methodology is same as FREAK [6] while only the sampling pattern being different.

\subsection{Algorithm}

\begin{figure}
  \centering
  \includegraphics[width=0.5\linewidth]{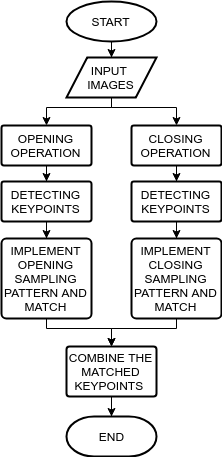}
  \caption{Flowchart}
  \label{fig:fc}
\end{figure}

Our method performs following steps as mentioned in Fig.~\ref{fig:fc}.

\begin{enumerate}
   \item Take images as input and implement morphological operations of opening and closing separately with kernel size $3\times3$. 
   \item Detect the keypoints in this images. 
   \item Apply the opening and closing sampling pattern around this keypoints.
   \item Match the keypoints obtained resulting in two different sets of matched keypoints.
   \item Combine this keypoints in the original images as shown in results to get the best set of matched keypoints.
\end{enumerate}

By combining the two sets of keypoints, we get more number of keypoints in the original set of images which in turn leads to better matching in many applications of computer vision.

\section{Observed results}

The required results are obtained by implementing it in python 3.5 in a Desktop PC with an intel i3 1.9 GHz processor and 8 GB of RAM. Opening and closing algorithms are implemented followed by description and matching using our approach by modifying \textit{$xfeatures2d$} in OpenCV 3.3.1 using python 3.5. Then FREAK algorithm is performed to compare results with our approach by considering three pairs of images as examples for evaluating results. Our approach is available on \textit{$https://github.com/himanshuvaghela/opencvcontrib$}.

Table.~\ref{fig:comp} shows computation time of each of the algorithms. Computation time to detect and match a keypoint in our approach is almost similar to that of FREAK. The characteristics of FREAK are inherited by our approach due to similar methodology but different sampling pattern.

\begin{figure}[p!]
  \centering
  \begin{subfigure}[b]{1\linewidth}
    \includegraphics[width=\linewidth]{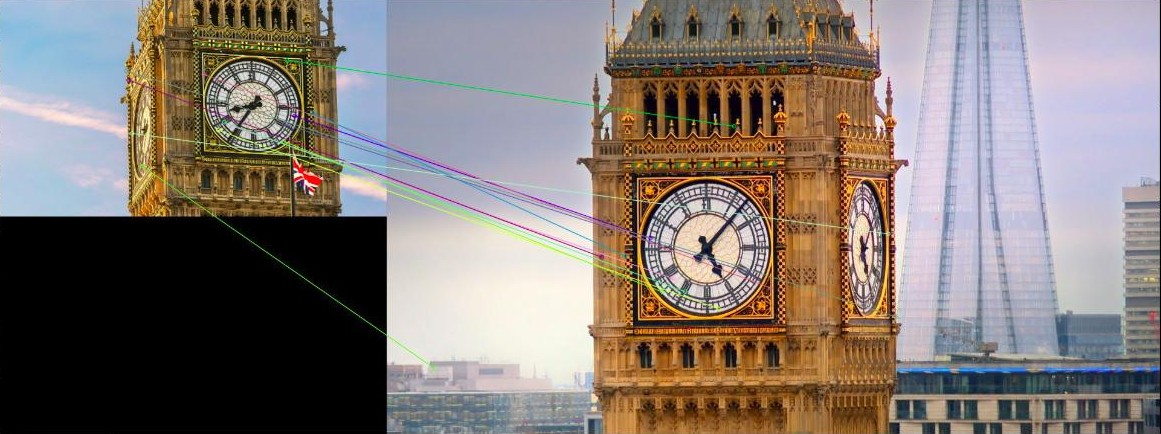}
    \caption{13 best FREAK keypoint matches.}
    \label{fig:e1a}
  \end{subfigure}
  \begin{subfigure}[b]{1\linewidth}
    \includegraphics[width=\linewidth]{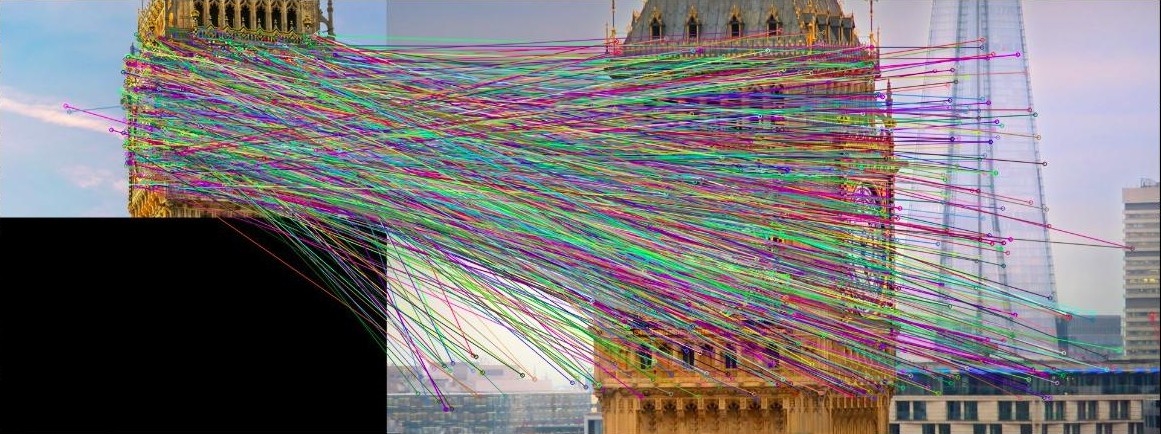}
    \caption{1498 total FREAK keypoints.}
    \label{fig:e1b}
  \end{subfigure}
  \begin{subfigure}[b]{1\linewidth}
    \includegraphics[width=\linewidth]{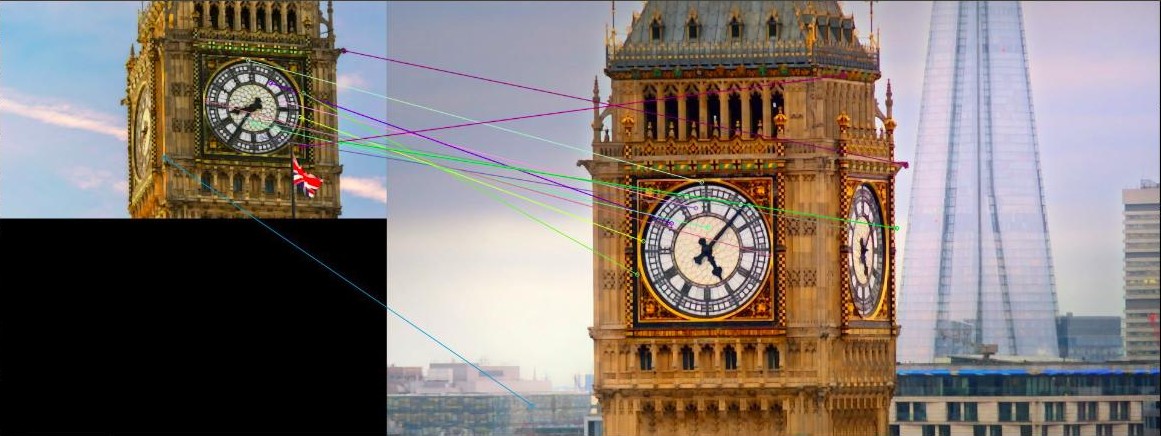}
    \caption{6 best keypoints by implementing opening sampling pattern in opened image.}
    \label{fig:e1c}
  \end{subfigure}
  \begin{subfigure}[b]{1\linewidth}
    \includegraphics[width=\linewidth]{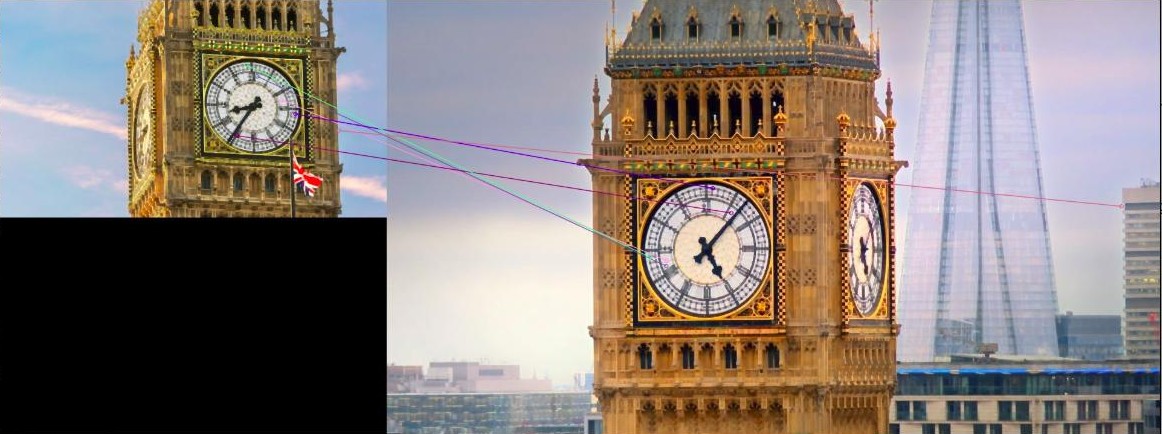}
    \caption{14 best keypoints by implementing closing sampling pattern in closed image.}
    \label{fig:e1d}
  \end{subfigure}
  \begin{subfigure}[b]{1\linewidth}
    \includegraphics[width=\linewidth]{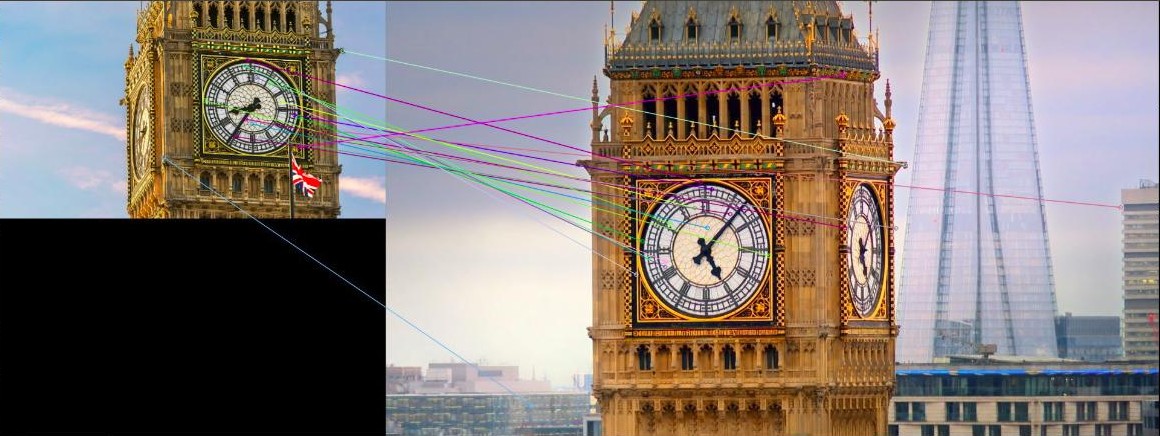}
    \caption{Combining total 20 best keypoints on original images.}
    \label{fig:e1e}
  \end{subfigure}
  \caption{Implementation on two images of Big Ben tower \cite{b12} and \cite{b13}}
\end{figure}

Consider an example of two photos of Big Ben tower in London \cite{b12} and \cite{b13}. Fig.~\ref{fig:e1a} shows 13 best matches of FREAK \cite{b6}. This best matches are determined by ratio of distances of closest neighbour to second closest as mentioned in Lowe’s SIFT \cite{b1}. In our case, ratio is greater than 7.5 which eliminates more than 80 $\%$ of false matches. Fig.~\ref{fig:e1b} shows 1498 total keypoint matches of FREAK. Fig.~\ref{fig:e1c} represents opening of two images and building descriptor using opening sampling pattern where 6 best keypoints are detected.  Fig.~\ref{fig:e1d} represents closing of two images and matching using closing sampling pattern where 14 best keypoints matches are detected. In Fig.~\ref{fig:e1e}, both of the opening and closing keypoints are combined and represented in original images forming a total of 20 best keypoints. By comparing  Fig.~\ref{fig:e1a} and Fig.~\ref{fig:e1e}, we can infer that extra keypoints are detected and better matching is done. There are few best matched keypoints in Fig.~\ref{fig:e1e} which  are not there in total FREAK keypoints in Fig.~\ref{fig:e1b}

\begin{figure}[p!]
  \centering
  \begin{subfigure}[b]{0.5\linewidth}
    \includegraphics[width=\linewidth]{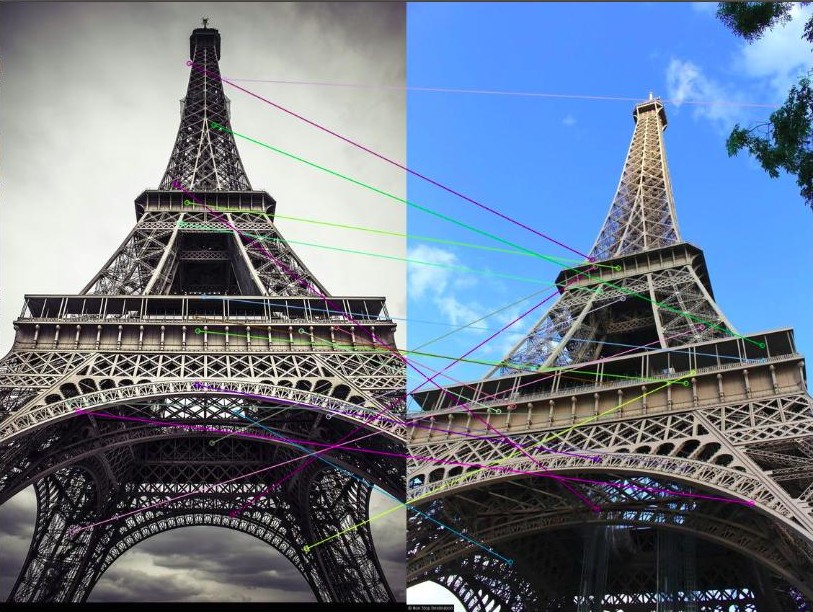}
    \caption{21 best FREAK keypoint matches.}
    \label{fig:e2a}
  \end{subfigure}
  \begin{subfigure}[b]{0.5\linewidth}
    \includegraphics[width=\linewidth]{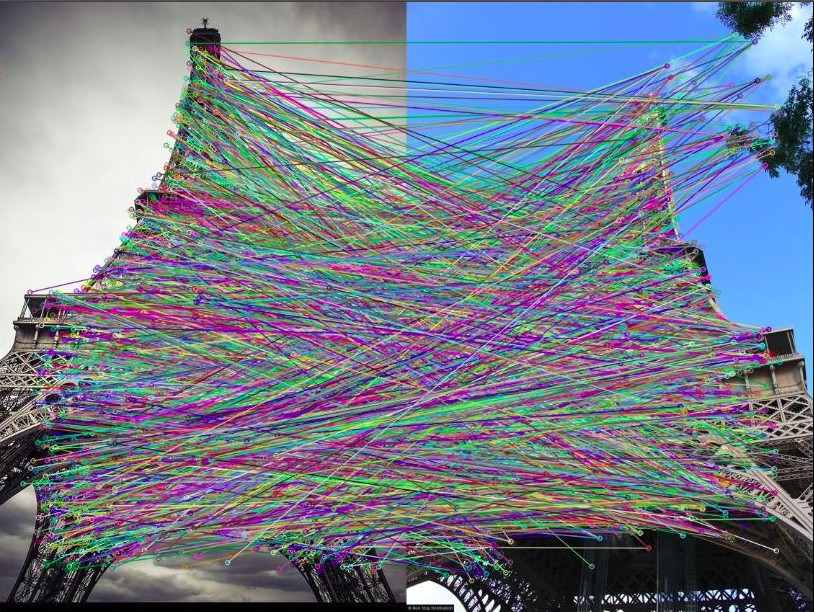}
    \caption{3897 total FREAK keypoints.}
    \label{fig:e2b}
  \end{subfigure}
  \begin{subfigure}[b]{0.5\linewidth}
    \includegraphics[width=\linewidth]{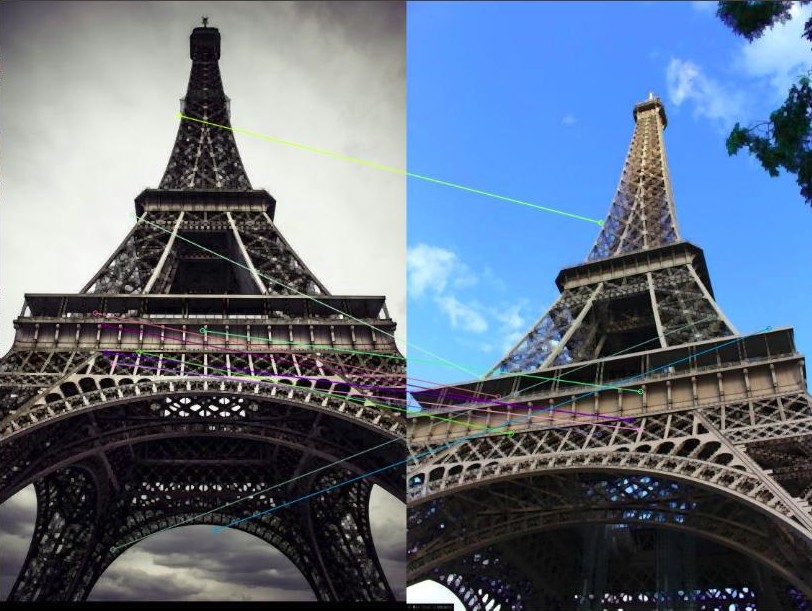}
    \caption{11 best keypoints by implementing opening sampling pattern in opened image.}
    \label{fig:e2c}
  \end{subfigure}
  \begin{subfigure}[b]{0.5\linewidth}
    \includegraphics[width=\linewidth]{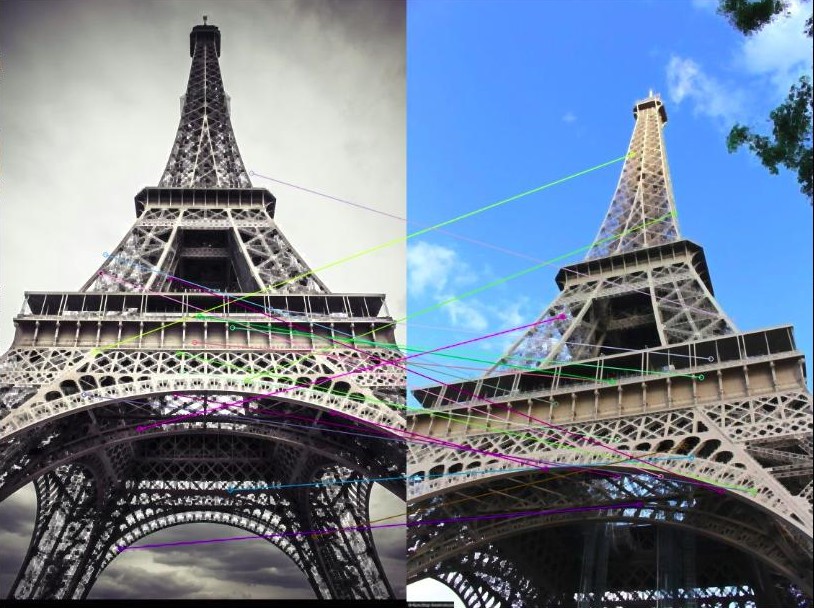}
    \caption{20 best keypoints by implementing closing sampling pattern in closed image.}
    \label{fig:e2d}
  \end{subfigure}
  \begin{subfigure}[b]{0.5\linewidth}
    \includegraphics[width=\linewidth]{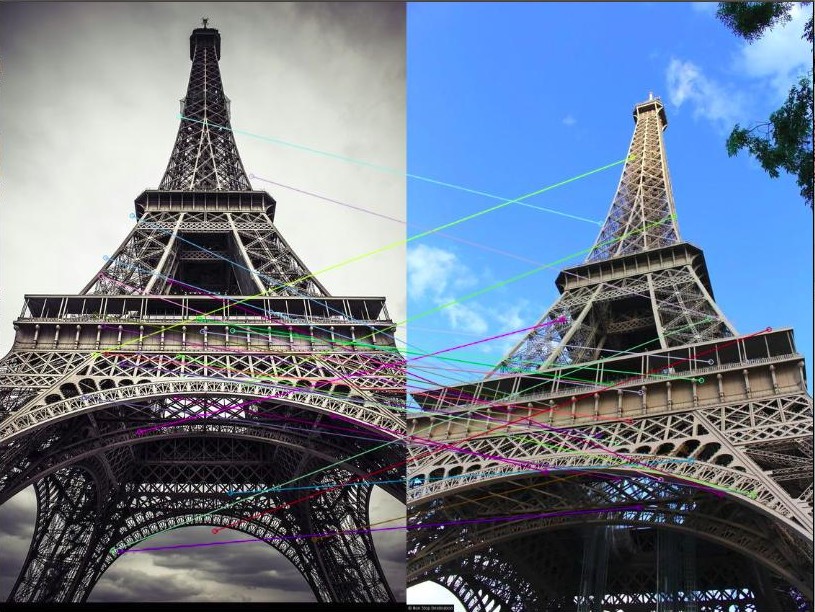}
    \caption{Combining total 31 best keypoints on original images.}
    \label{fig:e2e}
  \end{subfigure}
  \caption{Implementation on two images of Eiffel tower \cite{b14} and \cite{b15}}  
\end{figure}

\begin{figure}[p!]
  \centering
  \begin{subfigure}[b]{0.5\linewidth}
    \includegraphics[width=\linewidth]{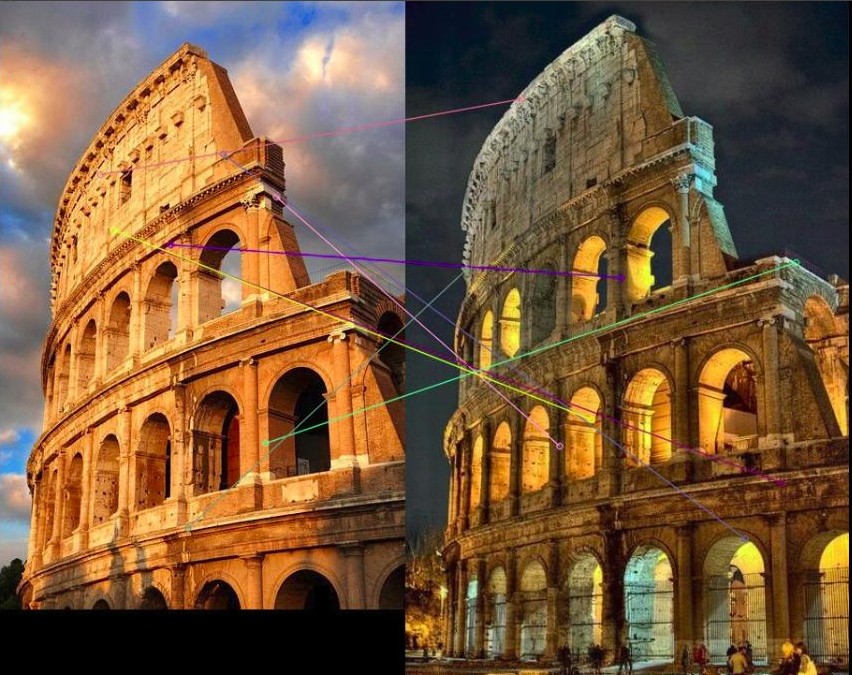}
    \caption{8 best FREAK keypoint matches.}
    \label{fig:e3a}
  \end{subfigure}
  \begin{subfigure}[b]{0.5\linewidth}
    \includegraphics[width=\linewidth]{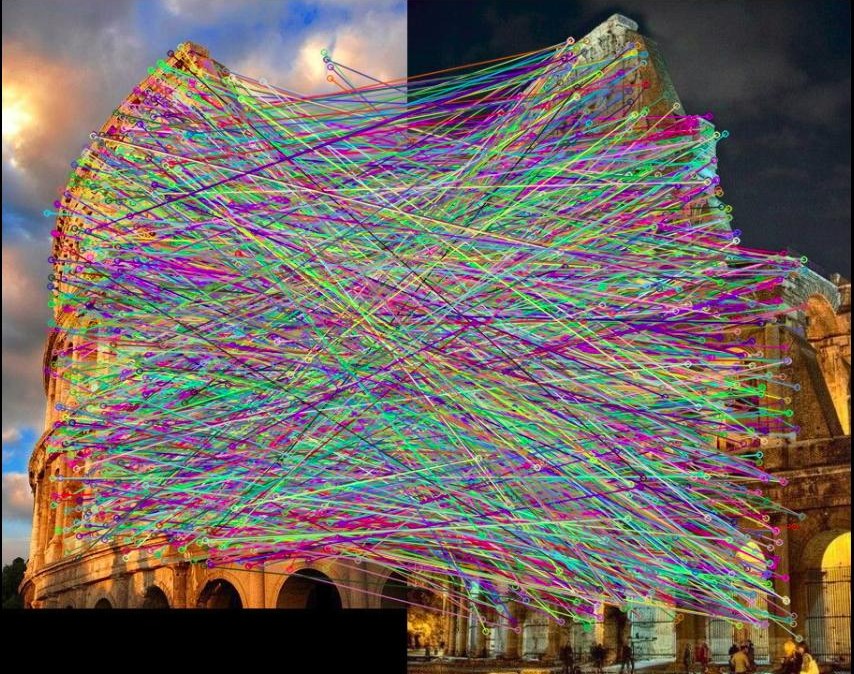}
    \caption{2027 total FREAK keypoints.}
    \label{fig:e3b}
  \end{subfigure}
  \begin{subfigure}[b]{0.5\linewidth}
    \includegraphics[width=\linewidth]{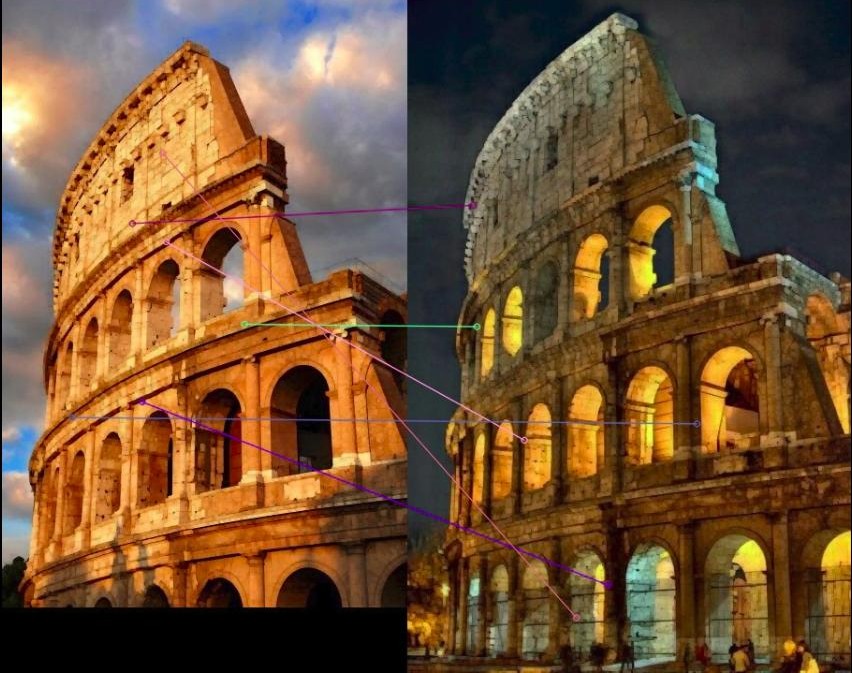}
    \caption{6 best keypoints by implementing opening sampling pattern in opened image.}
    \label{fig:e3c}
  \end{subfigure}
  \begin{subfigure}[b]{0.5\linewidth}
    \includegraphics[width=\linewidth]{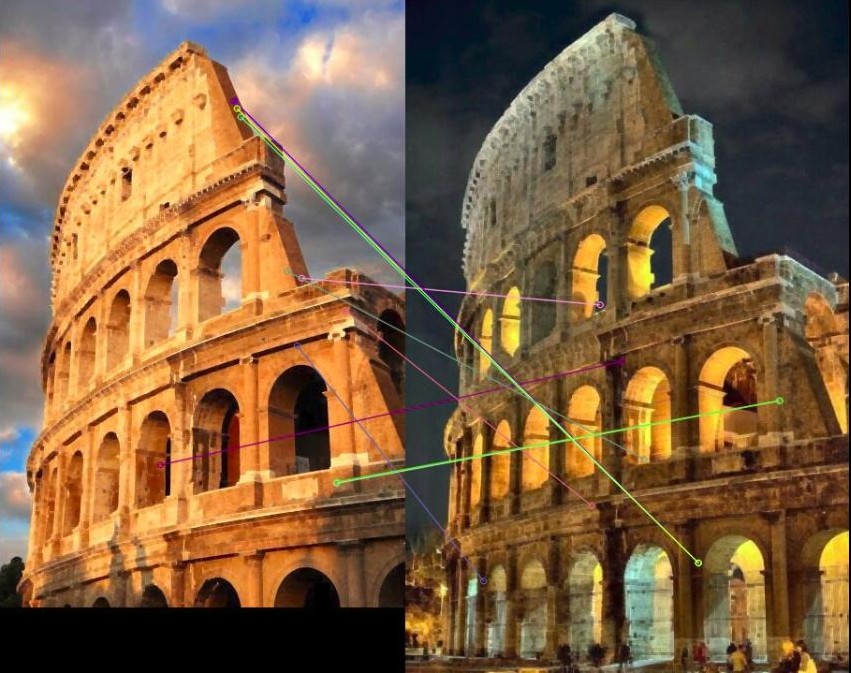}
    \caption{10 best keypoints by implementing closing sampling pattern in closed image.}
    \label{fig:e3d}
  \end{subfigure}
  \begin{subfigure}[b]{0.5\linewidth}
    \includegraphics[width=\linewidth]{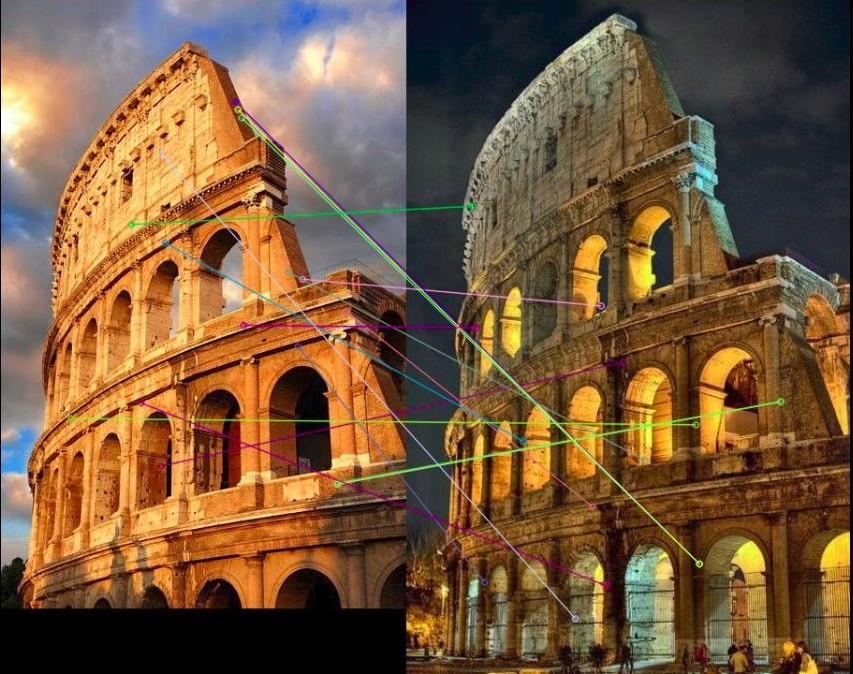}
    \caption{Combining total 16 best keypoints on original images.}
    \label{fig:e3e}
  \end{subfigure}
  \caption{Implementation on two images of Colosseum \cite{b16} and \cite{b17}} 
  
\end{figure}

In second example, two images of Eiffel tower are considered \cite{b14} and \cite{b15}. Fig.~\ref{fig:e2a} shows 21 best FREAK keypoint matches while Fig.~\ref{fig:e2e} shows 31 total best keypoints using our approach. In third example, two images of Colosseum \cite{b16} and \cite{b17} are considered. Fig.~\ref{fig:e3a} shows 8 best FREAK keypoint matches while Fig.~\ref{fig:e3e} shows 16 total best keypoints using our approach. 

It is observed that performing morphological operations like opening and closing on images increases the number of keypoints and a slight change in sampling pattern over FREAK improves the quality of keypoint matching providing a breakthrough in many computer vision applications. So, efficiency of results increases as we mimic the structure of human eye. This was proved earlier by FREAK and then by our approach where function of human pupil implemented through morphological operations eventually improved overall results.

\begin{table}[h!]
\caption{Computation time for 3 pair of images where approximately 2000 keypoints are detected per image.}
\begin{center}
 \begin{tabular}{|c|c|c|c|c|c|} 
 \hline
   Time per keypoint & SIFT & SURF & BRISK & FREAK & OUR \\  
   (ms) & & & & & METHOD \\
 \hline
 Description time & 0.0734 & 0.0631 & 0.0315 & 0.0280 & 0.0286 \\ 

 Matching time & 0.3716 & 0.3701 & 0.3788 & 0.1449 & 0.1164 \\
 \hline
\end{tabular}
\end{center}
\label{fig:comp}
\end{table}

\section{Conclusion}
In this paper, a binary descriptor inspired by FREAK sampling pattern and function of human pupil is presented. The field of computer vision can be further explored by mimicking human eye. The FREAK sampling pattern was inspired by distribution of ganglion cells in human retina. Our approach is inspired by the pupillary response of human eye. New keypoints were detected using this property by implementing selected morphological operations and accordingly improving the sampling pattern resulting in efficient matching of keypoints. Moreover, this concept of using morphological operations for improving the quality of images can be modified and used in many methods to improve the accuracy of keypoint matching and try to reduce computation time even further.

\end{document}